\documentclass[journal]{IEEEtran}
\pdfoutput=1

\usepackage{cite}

\ifCLASSINFOpdf
   \usepackage[pdftex]{graphicx}
\else
   \usepackage[dvips]{graphicx}
\fi
\usepackage[cmex10]{amsmath}

\DeclareMathOperator*{\argmin}{argmin}
\newcommand{\degree}{\ensuremath{^\circ}}

\usepackage{array}

\ifCLASSOPTIONcompsoc
  \usepackage[caption=false,font=normalsize,labelfont=sf,textfont=sf]{subfig}
\else
  \usepackage[caption=false,font=footnotesize]{subfig}
\fi
\usepackage{url}

\begin{document}
%
\title{Learning a time-dependent master saliency map \\from eye-tracking data in videos}
%
%
%

\author{Antoine~Coutrot
        and~Nathalie~Guyader
\thanks{A. Coutrot is with CoMPLEX, University College London, United Kingdom, e-mail: (see http://antoinecoutrot.magix.net/public/index.html).}
\thanks{N. Guyader is with Gipsa-lab, CNRS \& University of Grenoble-Alpes, France.}
}
\maketitle

\begin{abstract}
To predict the most salient regions of complex natural scenes, saliency models commonly compute several feature maps (contrast, orientation, motion...) and linearly combine them into a master saliency map. Since feature maps have different spatial distribution and amplitude dynamic ranges, determining their contributions to overall saliency remains an open problem. Most state-of-the-art models do not take time into account and give feature maps constant weights across the stimulus duration. However, visual exploration is a highly dynamic process shaped by many time-dependent factors. For instance, some systematic viewing patterns such as the center bias are known to dramatically vary across the time course of the exploration.
In this paper, we use maximum likelihood and shrinkage methods to dynamically and jointly learn feature map and systematic viewing pattern weights directly from eye-tracking data recorded on videos. We show that these weights systematically vary as a function of time, and heavily depend upon the semantic visual category of the videos being processed. Our fusion method allows taking these variations into account, and outperforms other state-of-the-art fusion schemes using constant weights over time. The code, videos and eye-tracking data we used for this study are available online\footnote{\url{http://antoinecoutrot.magix.net/public/research.html}}.
\end{abstract}

%
\IEEEpeerreviewmaketitle

\section{Introduction}
%
%
%
%
\IEEEPARstart{O}{ur} visual environment contains much more information than we are able to perceive at once. Attention allows us to select the most relevant parts of the visual scene and bring the high-resolution part of the retina, the fovea, onto them. The modeling of visual attention has been the topic of numerous studies in many different research fields, from neurosciences to computer vision. This interdisciplinary interest led to the publication of a large number of computational models of attention, with many different approaches, see \cite{Borji2013ha} for an exhaustive review. In fact, being able to predict the most salient regions in a scene leads to a wide range of applications, like image segmentation \cite{Mishra2009}, image quality assessment \cite{Ma2010}, image and video compression \cite{Guo2010}, image re-targeting \cite{Goferman2012ul}, video summarization \cite{Evangelopoulos2013fl}, object detection \cite{butko2009optimal} and recognition \cite{Han2010}, robot navigation \cite{Ruesch2008us}, human-robot interaction \cite{Zaraki2014ej}, retinal prostheses \cite{parikh2010}, tumours identification in X-rays images \cite{hong2003}.\\
\begin{figure*}[!t]
\centering
\includegraphics[width=18cm]{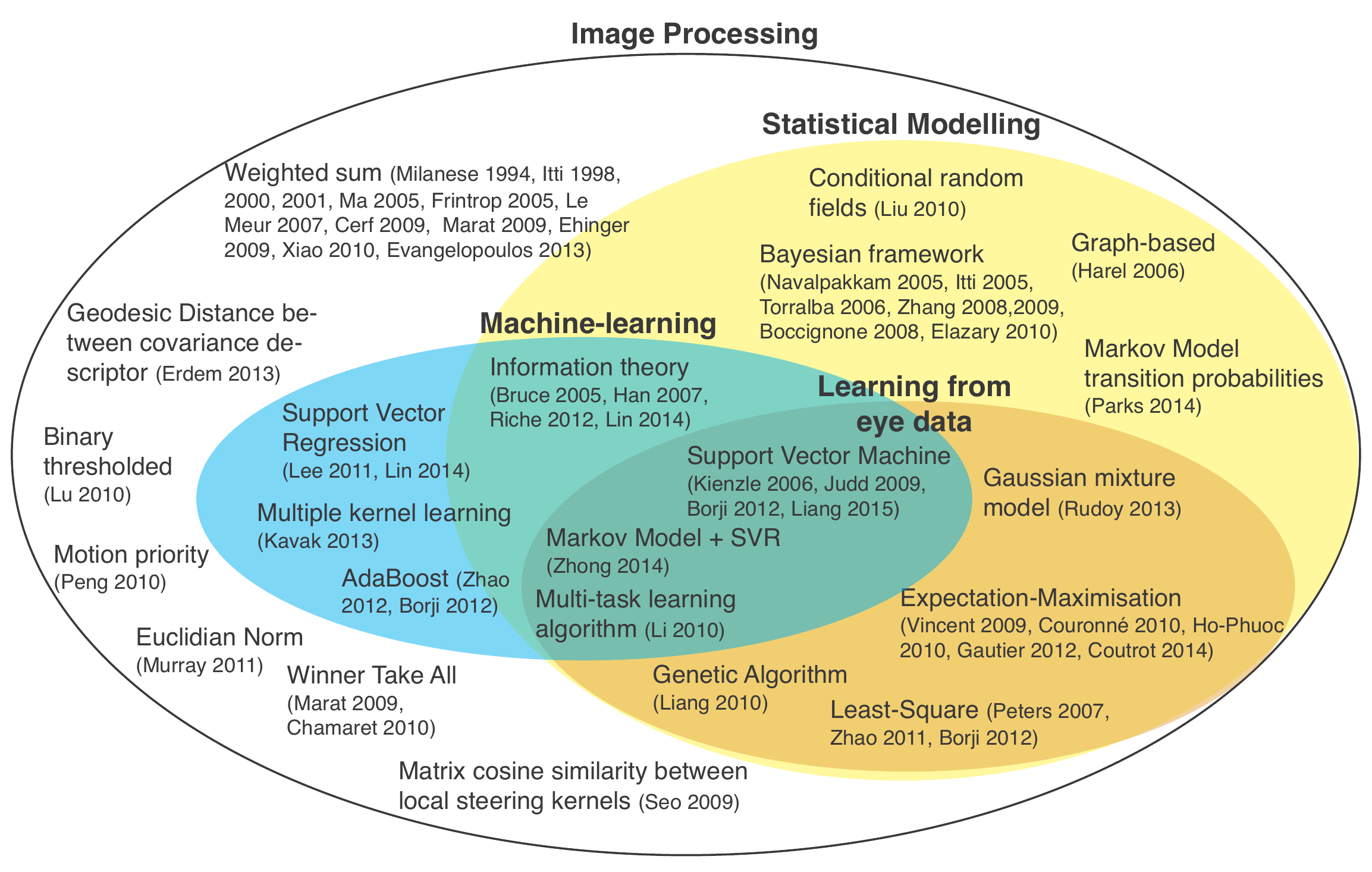}
\caption{State-of-the-art of feature map fusion schemes. They are classified in 4 families. 1- Image processing  methods, eventually relying on cognitive priors, 2- Machine-learning methods, 3- Statistical modeling methods, 4- Learning from eye-data methods. These families can intersect. From top to bottom and left to right: Milanese 1994 \cite{Milanese1994us}, Itti 1998, 2000, 2005, \cite{Itti1998tq, Itti2000te, Itti2005ur}, Ma 2005 \cite{Ma2005eg}, Frintrop 2005 \cite{Frintrop2005to}, LeMeur 2007 \cite{LeMeur2007dq}, Cerf 2009 \cite{Cerf2009jb}, Marat 2009 \cite{Marat2009fl}, Ehinger2009 \cite{Ehinger2009ji}, Xiao 2010 \cite{xiao2010video}, Evangelopoulos 2013 \cite{Evangelopoulos2013fl}, Erdem 2013 \cite{Erdem2013bg}, Lu 2010 \cite{Lu2010}, Peng 2009 \cite{peng2009}, Murray 2011 \cite{Murray2011ty}, Chamaret 2010 \cite{Chamaret2010wz}, Seo 2009 \cite{seo2009}, Lee 2011 \cite{lee2011}, Lin 2014 \cite{Lin2014db}, Kavak 2013 \cite{Kavak2013tg}, Zhao 2012 \cite{Zhao2012kx}, Borji 2012 \cite{Borji2012ux}, Bruce 2005 \cite{Bruce2005tt}, Han 2007 \cite{han2007}, Riche 2012 \cite{Riche2013fq}, Kienzle 2006 \cite{Kienzle2006uq}, Judd 2009 \cite{Judd2009fk}, Liang 2015 \cite{Liang2015go}, Liang 2010 \cite{Liang2010ud}, Liu 2010 \cite{Liu2010ba}, Navalpakkam 2005 \cite{navalpakkam2005}, Torralba 2006 \cite{Torralba2006tm}, Zhang 2008 \cite{Zhang2008ew}, Zhang 2009 \cite{zhang2009}, Boccignone 2008 \cite{boccignone2008}, Elazary 2010 \cite{Elazary2010ec}, Harel 2006 \cite{Harel2006vy}, Parks 2014 \cite{Parks2014fk}, Rudoy 2013 \cite{Rudoy2013tm}, Vincent 2009 \cite{Vincent2009hi}, Couronne 2010 \cite{Couronne2010vb}, Ho Phuoc 2010 \cite{HoPhuoc2010cm}, Gautier 2012 \cite{Gautier2012dm}, Coutrot 2014 \cite{Coutrot2014bx, Coutrot2014vy}, Peters 2007 \cite{Peters2007uv}, Zhao 2011 \cite{Zhao2011df}.}
\label{S_o_A}
\end{figure*}
Determining which location in a scene will capture attention - and hence the gaze of observers - requires finding the most salient subset of the input visual stimuli. Saliency relies on both stimulus-driven (bottom-up) and observer or task-related (top-down) features. Most visual attention models focus on bottom-up processes and are guided by the Feature Integration Theory (FIT) proposed by Treisman and Gelade \cite{Treisman1980tg}. They decompose a visual stimulus into several feature maps dedicated to specific visual features (such as orientations, spatial frequencies, intensity, luminance, contrast) \cite{Koch1985wv, Itti1998tq, LeMeur2007dq, Marat2009fl}. In each map, the spatial locations that locally differ from their surroundings are emphasized (conspicuity maps). Then, maps are combined into a master saliency map that points out regions the most likely to attract the visual attention, and the gaze, of observers. The visual features used in a model play a key role as they strongly affect its prediction performance. Over the years, authors refined their models by adding different features, such as center-bias \cite{Judd2009fk}, faces \cite{Marat2013dd}, text \cite{Cerf2009jb}, depth \cite{Gautier2012dm}, contextual information \cite{Torralba2006tm}, or sound \cite{Coutrot2014bx,Coutrot2014vy}. Other approaches, moving away from the FIT, have also been proposed. They often rely on machine-learning or statistical modeling techniques such as graphs \cite{Harel2006vy}, conditional random fields \cite{Liu2010ba}, Markov models \cite{Parks2014fk}, multiple kernel learning \cite{Kavak2013tg}, adaptive boosting \cite{Zhao2012kx}, Bayesian modeling \cite{Itti2005ur}, or information theory \cite{Lin2014db}.\\ 
Major problems arise when attempting to merge feature maps having different dynamic ranges, stemming from different visual dimensions. Various fusion schemes have been used. In Fig. \ref{S_o_A}, we propose a classification of these methods according to four main approaches: image processing (eventually relying on cognitive priors), statistical modeling, machine-learning and learning from eye data. Note that some methods might belong to two classes, like "statistical modeling" and "learning from eye data": statistical models can be used to learn the feature map weights that best fit eye-tracking data \cite{Rudoy2013tm, Vincent2009hi, Couronne2010vb, HoPhuoc2010cm, Gautier2012dm, Coutrot2014bx, Coutrot2014vy, Peters2007uv, Zhao2011df}. In the following, we focus on schemes involving a weighted linear combination of feature maps: 
\begin{equation}
S=\sum_{i=1}^{K} \beta_k M_k
\end{equation}
with S the master saliency map, $(M_k)_{k\in [1..K]}$ the maps of K features extracted from the image or video frame being processed, and $(\beta_k)_{k\in [1..K]}$ their associated weights. S and M are two-dimensional maps corresponding to the spatial dimensions of the visual scene.
Non-linear fusion schemes based on machine-learning techniques such as multiple kernel learning can be efficient, but they often suffer from being used as a "black box". On the other hand, linear combinations are easier to interpret: the greater the weight, the more important the corresponding feature map in the master saliency map. Moreover, such linear fusion is biologically plausible \cite{Nothdurft2000}. \\
In this linear context, we review two main approaches to determine the weights allocated to different feature maps. The first approach is based on priors (psychovisual or image-based), while the second is totally data-driven: the feature weights are learned to fit eye-tracking data.

\subsection{Computing weights via image processing and cognitive priors}
Itti \textit{et al.} proposed an efficient normalization scheme that had been taken up many times \cite{Itti1998tq,Itti2000te}. First, all feature maps are normalized to the same dynamic range. Second, to promote feature maps having a sparse distribution of saliency and to penalize the ones having a more uniform spatial distribution, feature map regions are weighted by their local maximum. Finally, feature maps are simply averaged.
Marat \textit{et al.} apply the same normalization, and use a fusion taking advantage of the characteristics of the static (luminance) and the dynamic (motion) saliency maps \cite{Marat2009fl}. Their weights are respectively equal to the maximum of the static saliency map and to the skewness of the dynamic saliency map.
A similar approach is adopted in \cite{Cerf2009jb}, where authors use a spatial competition mechanism based on the squared ratio of global maximum over average local maximum. This promotes feature maps with one conspicuous location to the detriment of maps presenting numerous conspicuous locations.  Then feature maps are simply averaged. In \cite{Frintrop2005to}, an analogous approach is proposed: each feature map is weighted by $1/\sqrt(m)$, with m the number of local maxima that exceed a given threshold.
In \cite{LeMeur2007dq}, authors propose a refinement of the normalization scheme introduced by Milanese \textit{et al.} in \cite{Milanese1994us}. First, the dynamic range of each feature map is normalized by using the theoretical maximum of the considered feature. The saliency map is obtained by a sum of maps representing the inter and intra-feature competition.\\
Other saliency models simply weight feature maps with coefficients determined by testing various values, and keep the best empirical weights \cite{Ma2005eg, Ehinger2009ji, xiao2010video,Evangelopoulos2013fl}.\\
Visual saliency models are often evaluated by comparing their outputs against the regions of the scene actually looked at by humans during an eye-tracking experiment. In some cases they can be very efficient, especially when attention is mostly driven by bottom-up processes. \\
\\
However, visual exploration is not only shaped by bottom-up visual features, but is also heavily determined by numerous viewing behavioral biases, or systematic tendencies \cite{Tatler2008uu}. Some models solely based on viewing biases, i.e. blind to any visual information from the input visual scene have been even shown to outperform state-of-the-art saliency models \cite{Tatler2009ea}. For instance, observers tend to look more at the center of a scene rather than at the edges; this tendency is known as the center bias \cite{Tatler2007hk, Tseng2009jn}. Some authors introduced this bias in their models through a static and centered 2D Gaussian map, leading to significant improvements \cite{Gautier2012dm, Marat2013dd}.
Many eye-tracking experiments have shown that the center bias is time-dependent: stronger at the beginning of an exploration than at the end \cite{Tatler2005vg, Coutrot2012vl}.
Moreover, several studies have pointed out that, in many contexts, top-down factors such as semantic or task clearly take the precedence over bottom-up factors to explain gaze behavior \cite{Nystrom2008vc,Rahman2015fw}. Visual exploration also relies on many individual characteristics such as personality or culture \cite{Chua2005vq,Risko2012dv}.
Thus, time-independent fusion schemes only considering the visual features of the input often have a hard time accounting for the multivariate and stochastic nature of human exploration strategies \cite{Tatler2011kk}.

\subsection{Learning weights from eye positions}
To solve this issue, a few authors proposed to build models in a supervised manner, by learning a direct mapping from feature maps to the eye positions of several observers. Feature maps could encompass classic low-level visual features as well as maps representing viewing behavioral tendencies such as the center bias.
The earliest learning-based approach was introduced by Peters \& Itti in \cite{Peters2007uv}. They used linear least square regression with constraints to learn the weights of feature maps from eye positions. Formally, let $(M_k)_{k\in [1..K]}$ be a set of K feature maps, $(\beta_k)_{k\in [1..K]}$ the corresponding vector of weights and Y an eye position density map, represented as the recorded fixations convolved with an isotropic Gaussian kernel. The least square (LS) method estimates the weights $\boldsymbol \beta$ by solving
\begin{equation}
\boldsymbol \beta^{LS} = \argmin_{\substack{\boldsymbol \beta}}\left\{(Y-\sum_{j=1}^{K} \beta_j M_{j})^2\right\} 
\label{eqLS}
\end{equation}
This method is repeated with success in \cite{Zhao2011df} and \cite{Borji2012ux}.
Another method to learn the weights $\boldsymbol \beta$ to linearly combine the visual feature maps is the Expectation-Maximisation algorithm \cite{Dempster1977ul}. It has first been applied in \cite{Vincent2009hi}, and taken over in \cite{Couronne2010vb,HoPhuoc2010cm,Gautier2012dm,Coutrot2014bx}. First, the eye position density map Y and the feature maps M are converted into probability density functions. After initializing the weights $\boldsymbol \beta$, the following steps are repeated until convergence. \textit{Expectation}: the current model (i.e. the current $\boldsymbol \beta$) is hypothesized to be correct, and the expectation of the model likelihood is computed (via the eye position data). \textit{Maximization}: $\boldsymbol \beta$ are updated to maximize the value found at the previous step.\\
To be exhaustive, let us mention some other models that do not use a weighted linear combination of feature maps, but that are still trained on eye-tracking data. In \cite{Kienzle2006uq,Judd2009fk,Borji2012ux,Liang2015go}, a saliency model is learnt from eye movements using a support vector machine (SVM). In \cite{Liang2010ud}, the authors refine a region-based attention model with eye-tracking data using a genetic algorithm. Finally, Zhong \textit{et al.} first use a Markov chain to model the relationship between the image feature and the saliency, and then train a support vector regression (SVR) from eye-tracking data to predict the transition probabilities of the Markov chain \cite{Zhong2014ii}.

\subsection{Contributions of the present study}
\begin{enumerate}
\item We provide the reader with a comprehensive review of visual feature fusion schemes in saliency models, exhaustive for gaze-based approaches.
\item So far, all these gaze-based saliency models have only been used with static images. Moreover, feature weights have generally been considered constant in time. Visual exploration is a highly dynamic process shaped by many time-dependent factors (e.g. center bias). To take these variations into account, we use a statistical shrinkage method (Least Absolute Shrinkage and Selection Operator, Lasso) to learn the weights of feature maps from eye data for a dynamic saliency model, i.e. for videos. 
\item For the first time, we demonstrate that the weights of the visual features depend both on time (beginning \textit{vs.} end of visual exploration) and on the semantic visual category of the video being processed. \textit{Dynamically} adapting the weights of the visual features across time allows to outperform state-of-the-art saliency models using constant feature weights.
\end{enumerate}


\section{Least Absolute Shrinkage and Selection Operator algorithm}
The different methods used in the literature to learn a master saliency map from eye-tracking data are enshrined within the same paradigm. Starting from predefined feature maps, they estimate the weights leading to the optimal master saliency map, i.e. the one that best fits actual eye-tracking data. Hence, the cornerstone of these methods is the estimation approach. Least square regression and Expectation-Maximization suffer from two flaws sharing the same root. The first one is prediction accuracy: least square estimates have low bias but large variance, especially when some feature maps are correlated. For instance, a high positive weight could be compensated by a high negative weight of the corresponding correlated feature. Second is interpretation: when a model is fitted with many features, it is difficult to seize the "big picture". In a nutshell, we would like to be able to sacrifice potentially irrelevant features to reduce the overall variance and improve the readability. This is precisely the advantage of Lasso over other learning methods such as least square regression or Expectation-Maximization \cite{Tibshirnani1996wb}.
Although widely spread in other fields such as genetic \cite{yi2008} or pattern recognition \cite{wright2010}, Lasso has never been used in the vision science society.
It is a shrinkage method. Given a model with many features, it allows selecting the most relevant ones and discarding the others, leading to a more interpretable and efficient model \cite{Hastie2009tz}. Lasso shrinks the feature weights $\boldsymbol \beta$ by imposing an L1-penalty on their size.
\begin{equation}
\boldsymbol \beta^{Lasso} = \argmin_{\substack{\boldsymbol \beta}}\left\{(Y-\sum_{j=1}^{K} \beta_j X_{j})^2 +  \lambda\sum_{j=1}^{K} |\beta_j|\right\} 
\label{eqLasso}
\end{equation}
with $\lambda>0$ a penalty parameter controlling the amount of shrinkage. If  $\lambda=0$, the Lasso algorithm corresponds to the least square estimate: all the feature maps are taken into account. If $\lambda \rightarrow \infty$, the weights $\boldsymbol \beta$ are shrunk toward zero, as well as the number of considered feature maps. Note that the feature maps M have to be normalized, Lasso being sensitive to the scale. Lasso is quite similar to the ridge regression \cite{Tikhonov1943}, just replacing the L2-penalty by a L1-penalty. The main advantage of L1 compared to L2 penalization is that L1 is even sparser: irrelevant feature maps are more effectively discarded \cite{Ng2004wz,Hastie2009tz}. Computing the Lasso solution is a quadratic programming problem, and different approaches have been proposed to solve the problem. Here, we adapted the code proposed in the \textit{Sparse Statistical Modeling toolbox} \cite{Sjostrand2012wf}, and made is available online\footnote{\url{http://antoinecoutrot.magix.net/public/code.html}}.  $\lambda$ is tuned to more or less penalize the feature map weights. A new vector $\boldsymbol \beta$ is determined at each step $\lambda+d\lambda$. The model (i.e. the $\boldsymbol \beta$ weights) having the weakest Bayesian Information Criterion (BIC) is chosen. BIC is a measure of the quality of a statistical model. To prevent overfitting, it takes into account both its likelihood and the number of parameters \cite{Schwarz1978}. For a model M and an observation Y, 
\begin{equation}
BIC(M | Y)=-2\log{L(M | Y)} + K\log{n}
\label{bic}
\end{equation}
with $L(M|Y)$ the likelihood, K the number of  feature maps, and n the number of points in Y. $\boldsymbol \beta$ weights are signed, and their sum are not necessarily unit, so for interpretability purposes, we normalized them between 0 and 1.

\section{Practical Application}
\begin{figure}[!t]
\centering
\includegraphics[width=10cm]{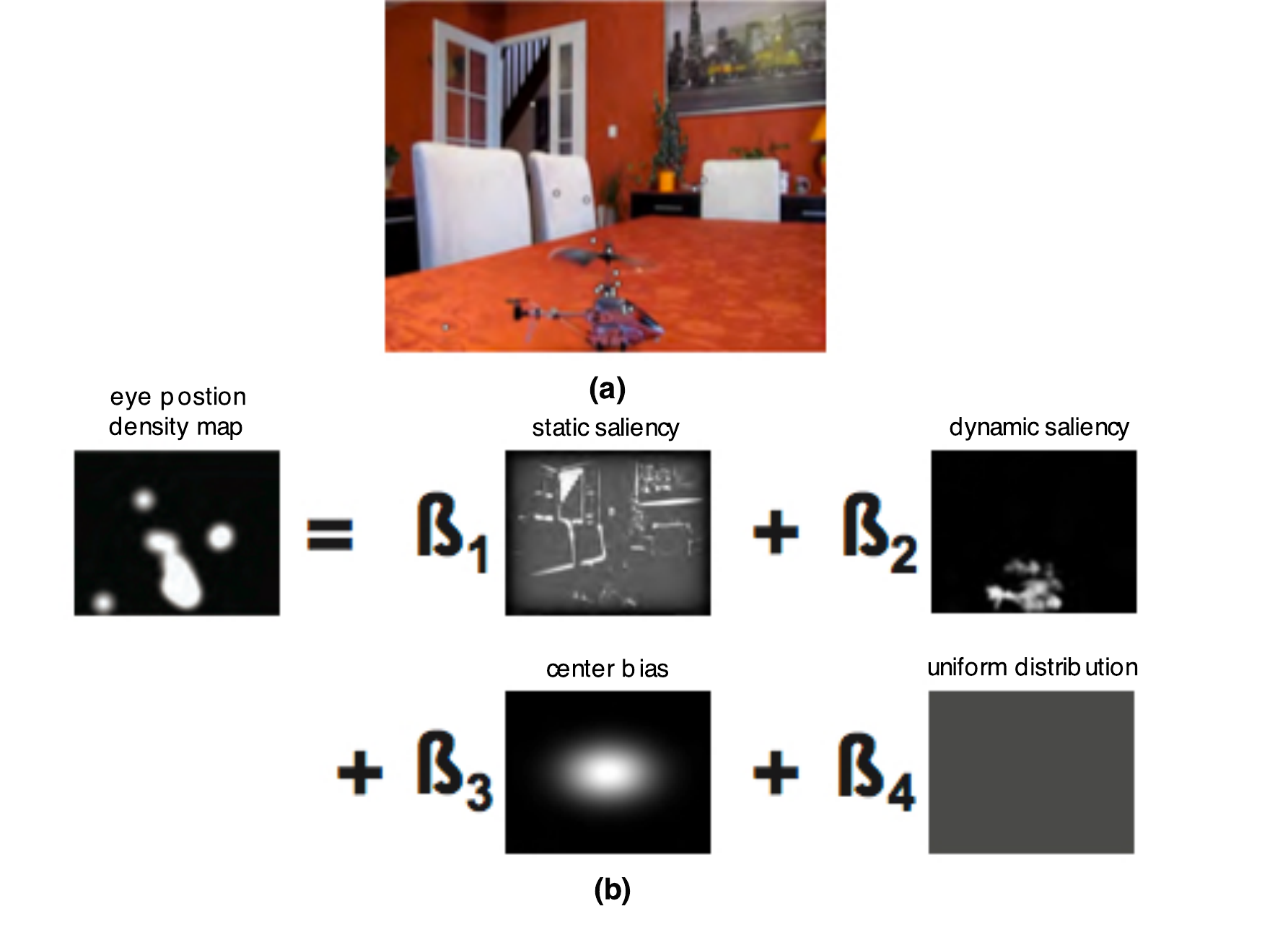}
\caption{(a) Frame extracted from a video belonging to One Moving Object category. The white and black dots represent the eye positions of participants.
(b) Modeling of the eye position density map (left) with a weighted linear combination of the static saliency map, the dynamic saliency map, a centre bias map, and a uniform distribution map.}
\label{Model1}
\end{figure}
\begin{figure}[!t]
\centering
\includegraphics[width=9.5cm]{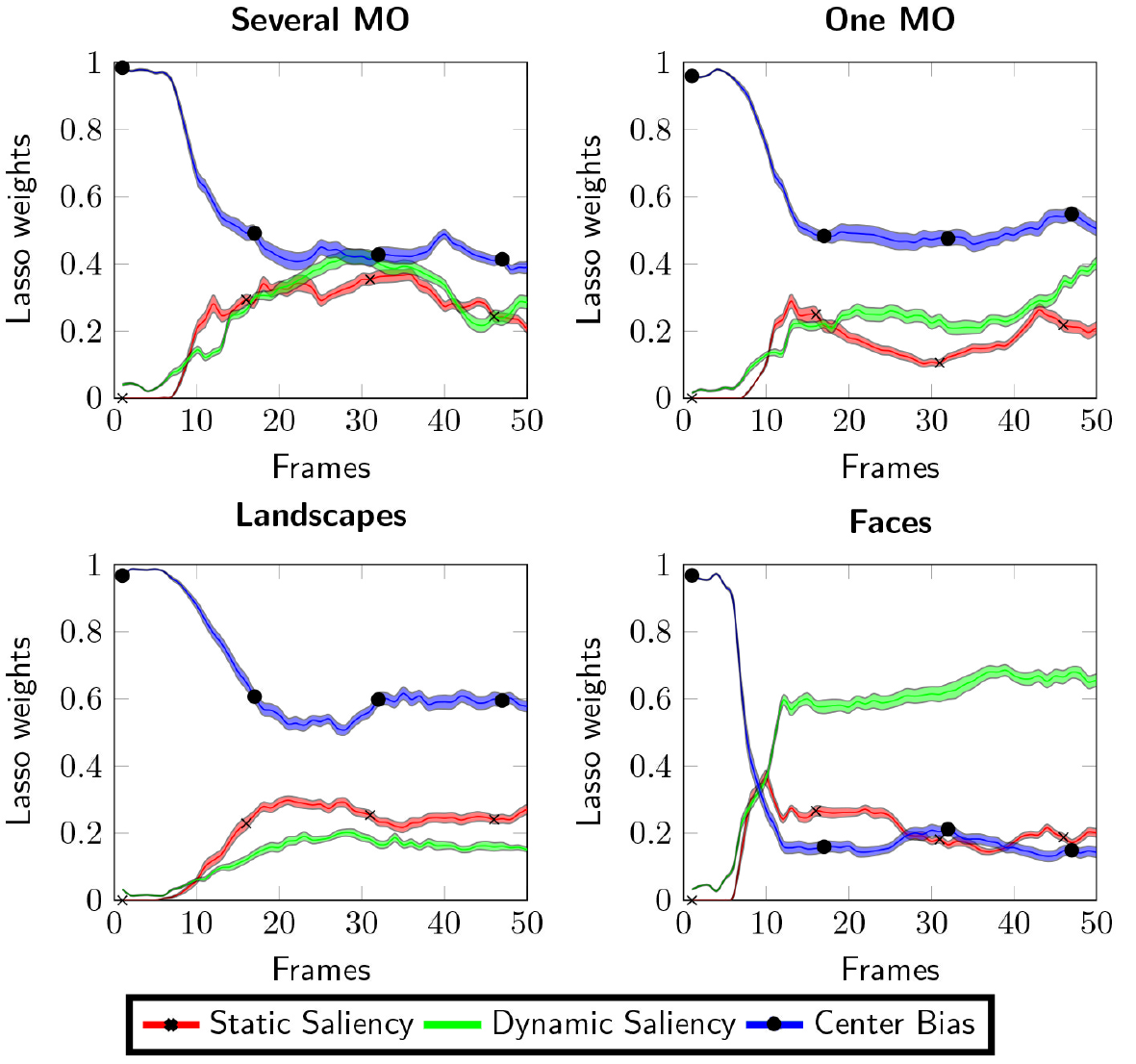}
\caption{Temporal evolution of the weights of static saliency map, dynamic saliency map and center bias map learned via the Lasso algorithm for the four visual categories of videos. The temporal evolution of the uniform distribution weight is not represented because equals to zero for all frames. Error bars represent s.e.m.}
\label{weightsModel1}
\end{figure}
We apply the Lasso algorithm to a video and eye-tracking database freely available online: \url{http://antoinecoutrot.magix.net/public/databases.html}
\subsection{Eye-tracking data}
The eye movements of 15 participants were recorded using an eye-tracker (Eyelink 1000, SR Research, Eyelink, Ottawa, Canada) with a sampling rate of 1000 Hz and a nominal spatial resolution of 0.01 degree of visual angle. We recorded the eye movements of the dominant eye in a monocular pupil-corneal reflection tracking mode. For more details, see \cite{Coutrot2014bx}.
\subsection{Visual material}
\begin{figure}[t]
\centering
\includegraphics[width=9.5cm]{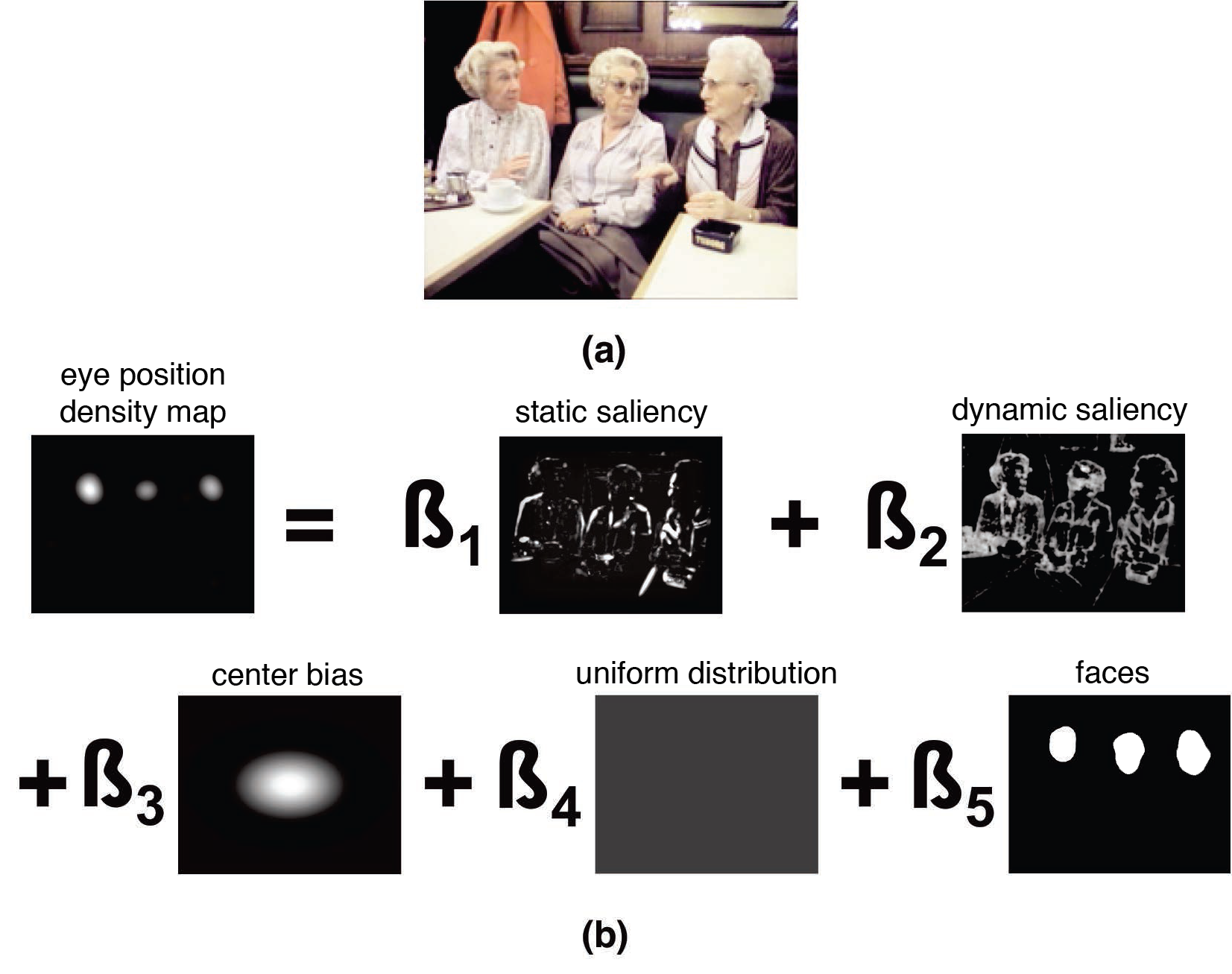}
\caption{(a) Frame extracted from a video belonging to Faces category. (b) Modeling of the eye position density map (left) with a weighted linear combination of static saliency map, dynamic saliency map, center-bias map, uniform map and a map with face masks.}
\label{Model2}
\end{figure}

\begin{figure}[t]
\centering
\includegraphics[width=9cm]{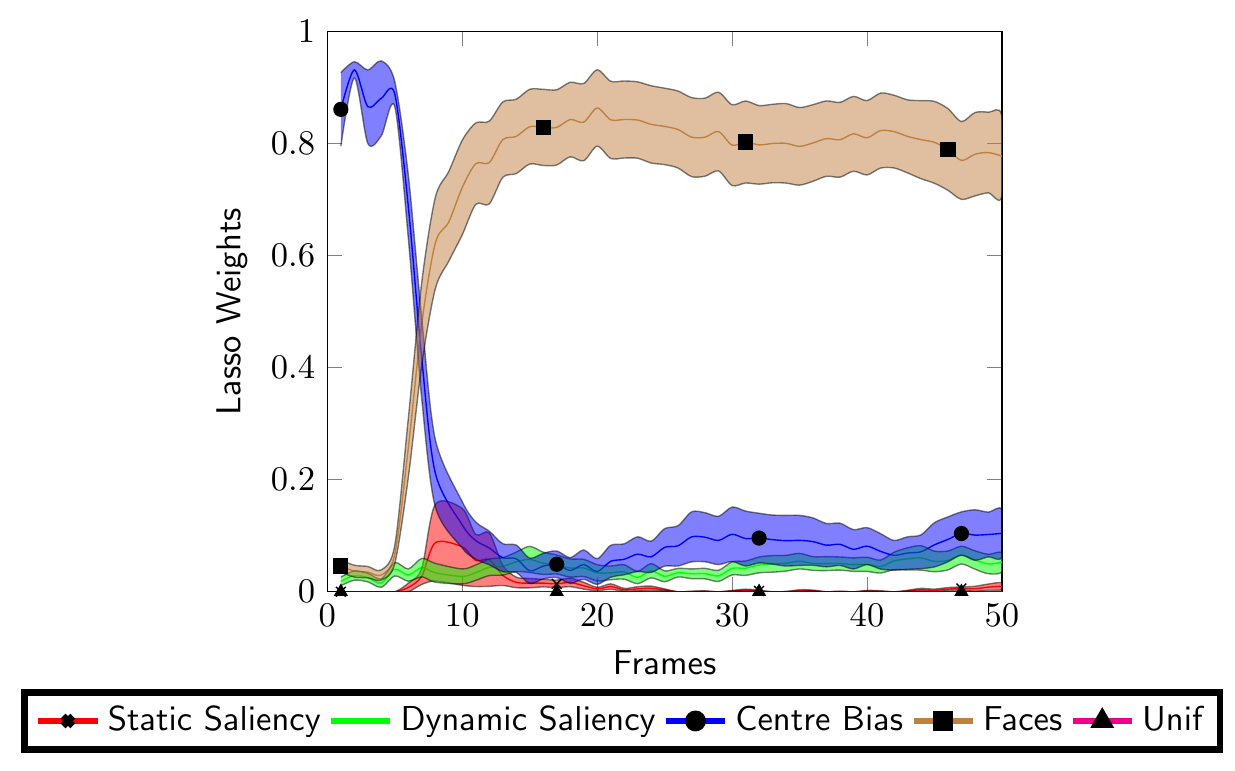}
\caption{Temporal evolution of the weights of static saliency map, dynamic saliency map, center bias map and face map, learned via the Lasso algorithm for videos belonging to Faces category. The temporal evolution of the uniform map weight is not represented because null. Error bars represent s.e.m.}
\label{weightsModel2}
\end{figure}

The videobase comprises 60 videos belonging to four visual categories: videos with faces, one moving object (one MO), several moving objects (several MO) and landscapes. Videos lasted between 10 and 31 seconds (720$\times$576 pixels, 25 fps), and were presented with their original monophonic soundtracks (Fs=48 kHz). Face videos present conversations between two to four people. We chose videos belonging to these categories since their regions of interest have very different spatial organization. For instance, observers' gaze would be clustered on speakers' face in Faces category, but would be more scattered in Landscapes category \cite{Coutrot2013up, Coutrot2014bx}.  
For each video frame, we computed the following feature maps.\\
\textbf{Eye position density map } A 2D Gaussian kernel (std = 1 degree of visual angle) is added to each of the 15 recorded eye positions.\\
\textbf{Static saliency map } We used the spatio-temporal saliency model proposed in \cite{Marat2009fl}. This biologically inspired model is based on luminance information. It processes the high spatial frequencies of the input to extract luminance orientation and frequency contrast through a bank of Gabor filters. Then, the filtered frames are normalized to strengthen the spatially distributed maxima. This map emphasizes the high luminance contrast.\\ 
\textbf{Dynamic saliency map } We used the dynamic pathway of the same model. First, camera motion compensation is performed to extract only the moving areas relative to the background. Then, through the assumption of luminance constancy between two successive frames, low spatial frequencies are used to extract moving areas. Finally, a temporal median filter is applied over five successive frames to remove potential noise from the dynamic saliency map. By giving the amplitude of the motion for each pixel, this map emphasizes moving areas.\\ 
\textbf{Centre bias } Most eye-tracking studies reported that observers tend to gaze more often at the center of the image than at the edges \cite{Tatler2007hk, Tseng2009jn}. As in \cite{Coutrot2014bx}, the center bias is modeled by a time-independent bi-dimensional Gaussian function centered at the screen center: $N(0,\Sigma)$, with $\Sigma = \bigl(\begin{smallmatrix} \sigma_x^2&0\\ 0&\sigma_y^2 \end{smallmatrix} \bigr)$ the covariance matrix and  $\sigma_x^2,  \sigma_y^2$ the variance. We chose $\sigma_x$ and $\sigma_y$ proportional to the frame size (28\degree x 22.5\degree). The results presented in this study were obtained with $\sigma_x = 2.3\degree$ and $\sigma_y = 1.9\degree$.\\
\textbf{Uniform map } All pixels have the same value (1 / (720$\times$576)). This map means that fixations might occur at all positions with equal probability. This feature is a catch-all hypothesis that stands for fixations that are poorly explained by other features. The lower the weight of this map is, the better the other features explain the eye fixation data.\\
Every map is then normalized as a 2D probability density function. As illustrated Fig. \ref{Model1}, we use the Lasso algorithm to learn for each frame the $\boldsymbol \beta$ weights that lead to the best fit of the eye position density map. Fig. \ref{weightsModel1}, we plot the temporal evolution of the feature weights across the first 50 frames (2 seconds). The uniform map weight is not represented as they all have been shrunk to zero. The weights are very different, both within and between visual categories. They all start by a sharp increase (static saliency, dynamic saliency) or decrease (centre bias) that lasts the first 15 frames (600 ms). Then, the weights plateau until the end of the video. The weight of static saliency stays small across the exploration, with no big difference between visual categories. The weight of dynamic saliency also is modest, except for Faces category. The weight of the centre bias is complementary to the one of the dynamic saliency, with rather high values, except for Faces visual category. \\
To better understand the singularity of Face maps, we added faces as a new feature for this visual category (Fig. \ref{Model2}). Face masks have been semi-automatically labelled using Sensarea. Sensarea is an authoring tool that automatically or semi-automatically performs spatio-temporal segmentation of video objects \cite{Bertolino2012vi}. We learned new $\boldsymbol \beta$ weights leading to the best fit of the eye position density map with the Lasso algorithm (Fig. \ref{weightsModel2}). After a short dominance of the center bias (similar to the one observed Fig. \ref{weightsModel1}), the face map weight clearly takes the precedence over the other maps. After that time, the uniform distribution weight is null, static saliency weight is negligible, and dynamic saliency and center bias weights are at the same level.
\begin{figure*}[t]
\centering
\includegraphics[width=12cm]{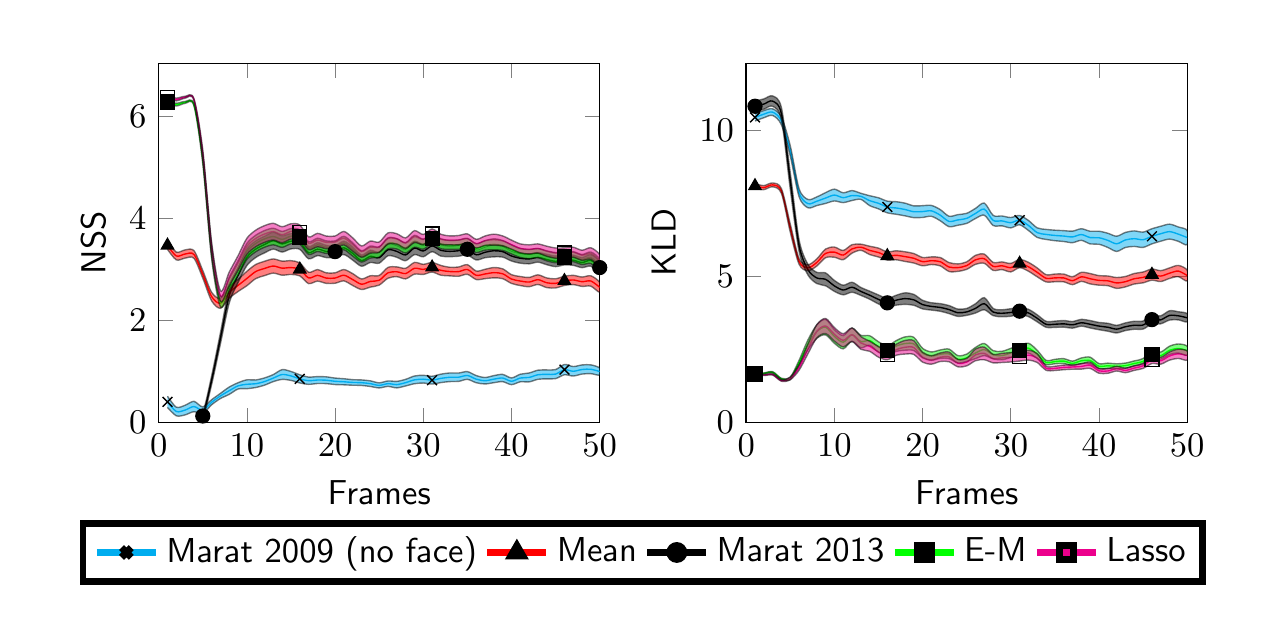}
\caption{Temporal evolution of the Normalized Scanpath Saliency (NSS, left) and the Kullback-Leibler Divergence (KLD, right) for 5 different fusion schemes. 1- Fusion of the static and dynamic saliency and center bias maps as proposed in Marat 2009 \cite{Marat2009fl}. 2- Simple mean of static and dynamic saliency, center bias and uniform maps. 3- Fusion of static and dynamic saliency, center bias and faces maps as proposed in Marat 2013 \cite{Marat2013dd}. 4- Weights of static and dynamic saliency, center bias and faces maps learned with the Expectation-Maximization (EM) algorithm and 5- with the Lasso algorithm. The greater the NSS and the lower the KLD, the better the model.  Error bars represent s.e.m.}
\label{metrics}
\end{figure*}
\subsection{Evaluation and comparison of fusion schemes}
We compared the performance of two time-dependent learning methods (Expectation-Maximization (EM) and Lasso) on Faces category with three other fusion schemes using the same feature maps. 
1) The fusion introduced in \cite{Marat2009fl}. This method only considers static saliency, dynamic saliency, and center bias maps, and does not take faces into account. After normalization, the authors multiply the static ($M_s$) and dynamic ($M_d$) saliency maps with the center bias map. Then, they use a fusion taking advantage of the characteristics of these maps:
\begin{equation}
S=\beta_s M_s + \beta_d M_d + \beta_s \beta_d M_s M_d
\end{equation}
with $\beta_s=max(M_s)$ and $\beta_d=skewness(M_d)$. Dynamic saliency maps with a high skewness correspond, in general, to maps with only small and compact areas. The term $\beta_s \beta_d$ gives more importance to the regions that are salient both in a static and dynamic way. 
2) The mean of every feature map: static saliency, dynamic saliency, center bias, uniform distribution and faces.
3) The fusion proposed in \cite{Marat2013dd}. This fusion scheme is similar to the one introduced in \cite{Marat2009fl}, but adds faces as a feature. A visual comparison of a time-dependent fusion (Lasso) with a time-independent fusion (Marat 2013) is available online.\\ 
For each frame, we compared the experimental eye position map with the master saliency map obtained for the four fusion schemes (Fig. \ref{metrics}). Not to evaluate the saliency maps with the same eye positions as the ones we used to estimate their corresponding weights, we followed a 'leave-one-out' approach. More precisely, the weights used to train the model for a given video originate from the average over the weights of every video but the one being processed.
We used two evaluation metrics: the Normalized Scanpath Saliency (NSS, \cite{Peters2005kq}) and the Kullback-Leibler Divergence (KLD, \cite{Kullback1951va}). We chose these two metrics because they are widely used for saliency model assessment \cite{LeMeur2013uk} and because they provide two different pictures of models' overall performance \cite{Riche2013wt}. The NSS acts like a z-score computed by comparing the master saliency map to the eye position density map. The KLD computes an overall dissimilarity between two distributions. The greater the NSS and the lower the KLD, the better the model. The results displayed Fig. 7 show a consistency between the two metrics: when the NSS of a fusion scheme is high, the corresponding KLD is low. We observe the same pattern as we did for the Lasso weights. There is a sharp increase or decrease that lasts the first 15 frames (600 ms). Then, the metrics plateau until the end of the video. We ran a two-way ANOVA with two within factors: the fusion scheme (Marat 2009 (no face), Mean, Marat 2013, EM and Lasso) and the period of visual exploration (1st and 2nd periods). We averaged NSS values over the first 15 frames (1st period, 600 ms) and from frame 16 to the end of the exploration (2nd period). We found an effect of fusion scheme (F(4,140)=161.4, $p<0.001$), period of visual exploration (F(1,140)=13.4, $p<0.001$ and of their interaction (F(4,140)=39.8, $p<0.001$). Simple main effect analysis showed that during the first period, all the fusion schemes are significantly different (all $p<0.001$), except EM and Lasso (p=0.98). EM and Lasso fusions are far above the other. During the second period, there is a significant difference between all fusion schemes, except between EM and Lasso (p=0.97), EM and Marat 2013 ($p=0.64$) and Lasso and Marat 2013 ($p=0.67$). We ran the same analysis on KLD scores. We found an effect of fusion scheme (F(4,140)=215.6, $p<0.001$), period of visual exploration (F(1,140)=92, $p<0.001$) and of their interaction (F(4,140)=24.8, $p<0.001$). Simple main effect analysis showed that during both periods, all the fusion schemes are significantly different (all $p<0.001$), except EM and Lasso (p=0.90 in 1st period, p=0.68 in 2nd period). Lasso and EM fusions are below the other.\\
Overall, both time-dependent learning fusions EM and Lasso lead to significantly better results with both metrics during the first period of exploration. During the second period, EM and Lasso fusions are still better with KLD, and perform on a similar level as Marat 2013 with NSS. Here, EM and Lasso perform quite similarly. This can be explained by the relatively small number of features involved in the model (static saliency, dynamic saliency, center bias, faces and uniform maps). Lasso would probably lead to a better combination for larger sets of features (e.g. 33 features in \cite{Judd2009fk}) by shrinking the weight of the less relevant ones to zero. 

\section{Discussion}
The focus of this paper is on time-dependent estimation of the weights of visual features extracted from dynamic visual scenes (videos). The weights are used to linearly combine low level visual feature and viewing bias maps into a master saliency map. The weights are learned frame by frame from actual eye-tracking data via the Lasso algorithm. We show that feature weights dramatically vary as a function of time and of the semantic visual category of the video.
\subsection{Influence of time}
Many studies have shown that gaze behavior strongly evolves across the time span of the exploration. This evolution is driven by both stimulus-dependent properties and task or observer-dependent factors. A simple way to quantify this evolution is by measuring the inter-observers eye position spatial dispersion across time. When this value is small, the observers tend to share a similar strategy; when it increases, the strategies are more diverse. Different eye-tracking studies converged toward the same temporal pattern: at the very beginning of the exploration, the inter-observers dispersion is small, then increases and reaches a plateau \cite{Tatler2005vg, Mital2010jo, Coutrot2012vl, Wang2012ki}. Moreover, if the stimulus is a video composed of several shots, the process is reset after each cut \cite{Coutrot2012vl}. The low dispersion at the beginning of the exploration can be explained by the center bias. Indeed, the center of the screen being an optimal location for early information processing of the scene, most observers start from there \cite{Tatler2007hk}.  Then, the observers adopt different exploration strategies, inducing an increase of the dispersion. When exploring static images, the dispersion keeps increasing \cite{Parkhurst2002vo, Tatler2005vg} while in dynamic scenes, the constant appearance of new information promotes bottom-up influences at the expense of top-down strategies. This induces a stable consistency between participants over time \cite{Carmi2006gk, Marat2009fl, Coutrot2012vl}. 
In this paper, we found the same temporal pattern in the weights of the feature maps. A short dominance of the center bias followed by a sharp increase (static saliency, dynamic saliency, faces) or decrease (centre bias) that lasts the first 15 frames (600 ms). Then, the weights plateau until the end of the video. Hence, one of the advantages of learning the weights of feature maps via actual eye-tracking data is that it allows capturing some oculomotor systematic tendencies, such as the center bias. One could argue that this interest is limited since it only concerns the first 600 ms of the exploration. However, an analysis of 160 Hollywood-style films released from 1935 to 2010 shows a linear decrease of mean shot duration \cite{Cutting2011bh}. Over 75 years, average shot durations have declined from about 15 to only 3.5 s. Since the center bias dominance is reset after each cut, it concerns almost 20\% of the movie.
\subsection{Influence of the semantic visual category of the video}
Our results show that feature weights also heavily rely on the semantic visual category of the video being processed. For instance, weights of dynamic saliency maps are globally higher in Several Moving Objects category than in Landscapes category, where motion is almost absent. The only exception concerns the weights of static saliency maps, which remain quite low for every visual category. Previous studies have shown that low-level static saliency predicts gaze position very poorly \cite{Carmi2006gk,Mital2010jo,Tatler2011kk}. In \cite{Tatler2009ea}, Tatler \& Vincent even showed that a model solely based on oculomotor biases, and thus blind to current visual information, outperforms popular low-level saliency models. Our results are in line with this finding: in every visual category, weights of static saliency maps are small compared to the ones of center bias. When the visual scene involves faces, the situation is different. The weights of dynamic saliency maps are much higher, at the expense of the weights of center bias. However, adding faces as a feature for this category show that the face regions explain around 80\% of the recorded eye positions. These two models (with and without considering face as a feature) show that face regions are correlated with motion. This can be understood under the light of the audiovisual speech literature, where many studies have pointed out the importance of gestures (lip motion, facial dynamic expression, head movements) in speech perception \cite{Summerfield1987, Schwartz1998, Bailly2012, Vo2012go}. 

\section{Conclusion}
When exploring a complex natural scene, gaze can be driven by a multitude of factors including bottom-up features, top-down features, and viewing biases. In this paper we show that algorithms such as Lasso or Expectation-Maximization can be used to jointly estimate the variations of the weight of all these features across time and semantic visual categories directly from eye-tracking data. This allows to tune the feature weights across the time course of the exploration. By doing so, our method outperforms other state-of-the-art fusion strategies based on the same visual features. 
Since many studies have shown systematic variations in gaze behavior patterns between different groups of people (e.g. based on culture, age, gender...), our method could be used to optimally tune a saliency model for a specific population. For instance, it has been shown that while viewing photographs with a focal object on a complex background, Americans fixate more on focal objects than do the Chinese, who make more saccades to the background \cite{Chua2005vq}. Hence, training a model with Chinese's or Americans' eye data would certainly lead to different feature weights. In the same vein, males have been shown to be less exploratory than females, especially when looking at faces \cite{Coutrot2016kj}. This approach can be compared with saccadic models \cite{LeMeur2015io,LeMeur2016jma}. Instead of outputting saliency maps, saccadic models generate plausible visual scanpaths, i.e. the actual sequence of fixations and saccades an observer would do while exploring stimuli. Like our Lasso-based learning model, saccadic models can be tailored for specific populations with specific gaze patterns (e.g. younger vs. older observers). Another application of our method is related to saliency aggregation. Recent studies have shown that combining saliency maps from different models can outperform any saliency model taken alone \cite{LeMeur2014ub, Danko2015jj}. Applying the Lasso algorithm to find the best saliency model combination could significantly enhance the accuracy and the robustness of the prediction.

\ifCLASSOPTIONcaptionsoff
  \newpage
\fi



\bibliographystyle{./IEEEtran}
\bibliography{./biblioIEEE}
%
%
%

%

\begin{IEEEbiography}{Antoine Coutrot}
Biography text here.
\end{IEEEbiography}

\begin{IEEEbiography}{Nathalie Guyader}
Biography text here.
\end{IEEEbiography}






\end{document}